\documentclass[]{spie}  

\usepackage{amsmath,amsfonts,amssymb}
\usepackage[colorlinks=true, allcolors=blue]{hyperref}

\usepackage{graphicx}

\title{Unsupervised Pathology Image Segmentation \\
Using Representation Learning with Spherical K-means}

\author[a]{Takayasu Moriya}
\author[a]{Holger R. Roth}
\author[b]{Shota Nakamura}
\author[c]{Hirohisa Oda}
\author[c]{Kai Nagara}
\author[a]{\\Masahiro Oda}
\author[a]{Kensaku Mori}
\affil[a]{Graduate School of Informatics, Nagoya University}
\affil[b]{Nagoya University Graduate School of Medicine}
\affil[c]{Graduate School of Information Science, Nagoya University}

\usepackage{amsmath}
\usepackage{amsfonts}
\usepackage{arydshln} 
\usepackage{subfig}

\newcommand{\argmax}{\mathop{\rm arg~max}\limits}
\newcommand{\argmin}{\mathop{\rm arg~min}\limits}

\begin{document}
\maketitle

\begin{abstract}

This paper presents a novel method for unsupervised segmentation of pathology images. Staging of lung cancer is a major factor of prognosis. Measuring the maximum dimensions of the invasive component in a pathology images is an essential task. Therefore, image segmentation methods for visualizing the extent of invasive and noninvasive components on pathology images could support pathological examination. However, it is challenging for most of the recent segmentation methods that rely on supervised learning to cope with unlabeled pathology images. In this paper, we propose a unified approach to unsupervised representation learning and clustering for pathology image segmentation. Our method consists of two phases. In the first phase, we learn feature representations of training patches from a target image using the spherical $k$-means. The purpose of this phase is to obtain cluster centroids which could be used as filters for feature extraction. In the second phase, we apply conventional $k$-means to the representations extracted by the centroids and then project cluster labels to the target images. We evaluated our methods on pathology images of lung cancer specimen. Our experiments showed that the proposed method outperforms traditional $k$-means segmentation and the multithreshold Otsu method both quantitatively and qualitatively with an improved normalized mutual information (NMI) score of 0.626 compared to 0.168 and 0.167, respectively. Furthermore, we found that the centroids can be applied to the segmentation of other slices from the same sample.

\end{abstract}

\keywords{Segmentation, Pathology, Representation Learning, Unsupervised Learning}

\section{Purpose}
\label{sec:intro}  
\begin{figure}[h]
  \centering
  \includegraphics[keepaspectratio, scale=0.5]{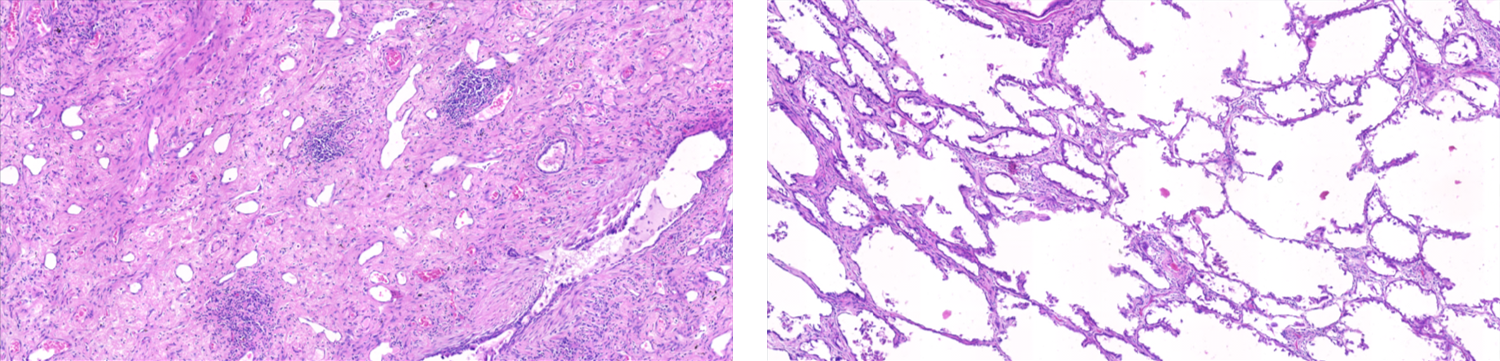}
  \caption{
  \textbf{Left:} invasive component in lung adenocarcinoma.
  The largest diameter of the invasive component is significant prognostic factor for lung cancer.
  \textbf{Right:} noninvasive component in lung adenocarcinoma.
  On pathology images, noninvasive components are observed as lepidic features.
  }
  \label{fig:img}
\end{figure}

The purpose of our study is to develop a novel unsupervised segmentation method of pathology images.
Staging of lung cancer is a major factor of prognosis.
Measuring the maximum dimensions of the invasive component in a pathology images (see Fig. \ref{fig:img}) is an essential task \cite{detterbeck2017eighth}.
Furthermore, measuring the maximum dimensions of the total tumor including the noninvasive components is also important \cite{travis2016iaslc}.
Therefore, segmentation methods for visualizing invasive and noninvasive components on pathology images could assist the pathological examination.
In our study,
we investigated whether representations learned by an unsupervised method aid in the segmentation of pathology images.
Research on unsupervised segmentation methods, especially for pathology images, is very promising because of the difficulty of obtaining manual annotations.

Our main contribution is to combine unsupervised representation learning with conventional clustering for pathology image segmentation.
As an unsupervised representation learning, we adopt spherical $k$-means \cite{dhillon2001concept}.
Spherical $k$-means training is much faster and easier implemented than CNN-based training.
For clustering, we adopt conventional $k$-means \cite{macqueen1967some} .
To our knowledge, our method is the first to employ spherical $k$-means to learn feature representations for unsupervised segmentation.

%
%

\section{METHOD}
The proposed segmentation method consists of two phases: (1) unsupervised learning of feature representations using spherical $k$-means and (2) segmentation by applying conventional $k$-means  to feature representations.
In phase (1), we conduct spherical $k$-means in order to learn the feature representations of image patches randomly extracted from an unlabeled image.
The purpose of this phase is to obtain centroids that can transform image patches to discriminative feature representations.
In phase (2), we use conventional $k$-means to assign labels to the representations extracted by the centroids on the full image.

\subsection{Representation Learning}
\label{sec:title}
It is known that spherical $k$-means can be used as representation learning method \cite{coates2012learning}.
Given a set of $N$ image patches $X = \{ {x}_{1} \dots x_{N} \}$,
spherical $k$-means aims to find optimal centroids $\hat{\mathcal{D}}$ by dividing data points into $K$ clusters according to:
\begin{eqnarray}
  (\hat{\mathcal{D}}, \hat{z})  =  \argmin_{{\mathcal{D}}, {z}}\displaystyle\sum_{i}
  \|{\mathcal{D}}{z}_{i} - {x}_{i}\|^{2}
\label{eq:s_kmeans}
\end{eqnarray}
where ${z}_{i} \in \mathbb{R}^{K}$ is a representation of ${x}_{i}$, called "code vector".
Centroids ${\mathcal{D}}$ and a code vector ${z}_{i}$ have constraints respectively to make it possible to reconstruct $x_{i}$ when given ${\mathcal{D}}$ and ${z}_{i}$ .
A code vector ${z}_{i}$ need to meets $\|{z}_{i}\|_{0} \le 1$ to have at most a single non-zero entry.
Centroids ${\mathcal{D}}$ need to satisfy the constraint that each column of ${\mathcal{D}}$ have unit length.
In order to accomplish Equation \ref{eq:s_kmeans},
we alternately optimize ${z}_{i}$ and ${\mathcal{D}}$
as follows:
\begin{eqnarray}
  \hat{{z}^{(i)}_{j}} &= &
  \begin{cases}
    {\mathcal{D}}_{j}^{T}{x}_{i}, & \text{if }j = \argmax_{l} |{\mathcal{D}}_{l}^{T}{x}_{i}| \\
    0 & \text{otherwise}
  \end{cases}
  \forall j,i \\
  {\mathcal{D}}' &=  &{X}{Z}^{T} + {\mathcal{D}} \\
  \hat{\mathcal{D}}_{j} &= &{{\mathcal{D}}_{j}}' /\|{{\mathcal{D}}_{j}}'\|, \forall j
\label{eq:s_kmeans_iter}
\end{eqnarray}
where ${Z}$ is the matrix whose columns are the code vectors ${z}_{i}$.
Optimal centroids could be used as filters which extract features\cite{coates2012learning}.
Spherical $k$-means can be rapidly executed so that $K$ can be set to a large value (e.g., $K = 1000$).
Thus, we can obtain large centroids and learn a large number of features.

\subsection{Pre-processing}
We extract training patches $X$ from an unlabeled target image by randomly cropping $N$ sub-images of $p\times p$ pixels.
Note that we should carefully choose the patch size because $k$-means feature learning is significantly sensitive to the dimensionality of the input data.
%
Note that we need to set a proper threshold in order not to include patches from the background.
After extracting training patches, we normalize the brightness and contrast of each patch.
While Coates \& Ng\cite{coates2012learning} use mean and variance of each patch $x_i$, we instead use mean and variance of entire dataset $X$.
The previous application of spherical $k$-means aims to classify test images which are independent of each other.
In contrast, our method aims to cluster patches from the same image which are not independent.
By using global mean and variance, we can retain the relative intensities among patches.
%
A previous study has shown that correlation has a bad effect on image recognition experiments \cite{coates2010analysis}.
After normalization, we apply ZCA whitening transform \cite{bell1997independent} to the normalized patches in order to decorrelate them.

\subsection{Segmentation}
\begin{figure}[tb]
  \centering
  \includegraphics[keepaspectratio, scale=0.485]{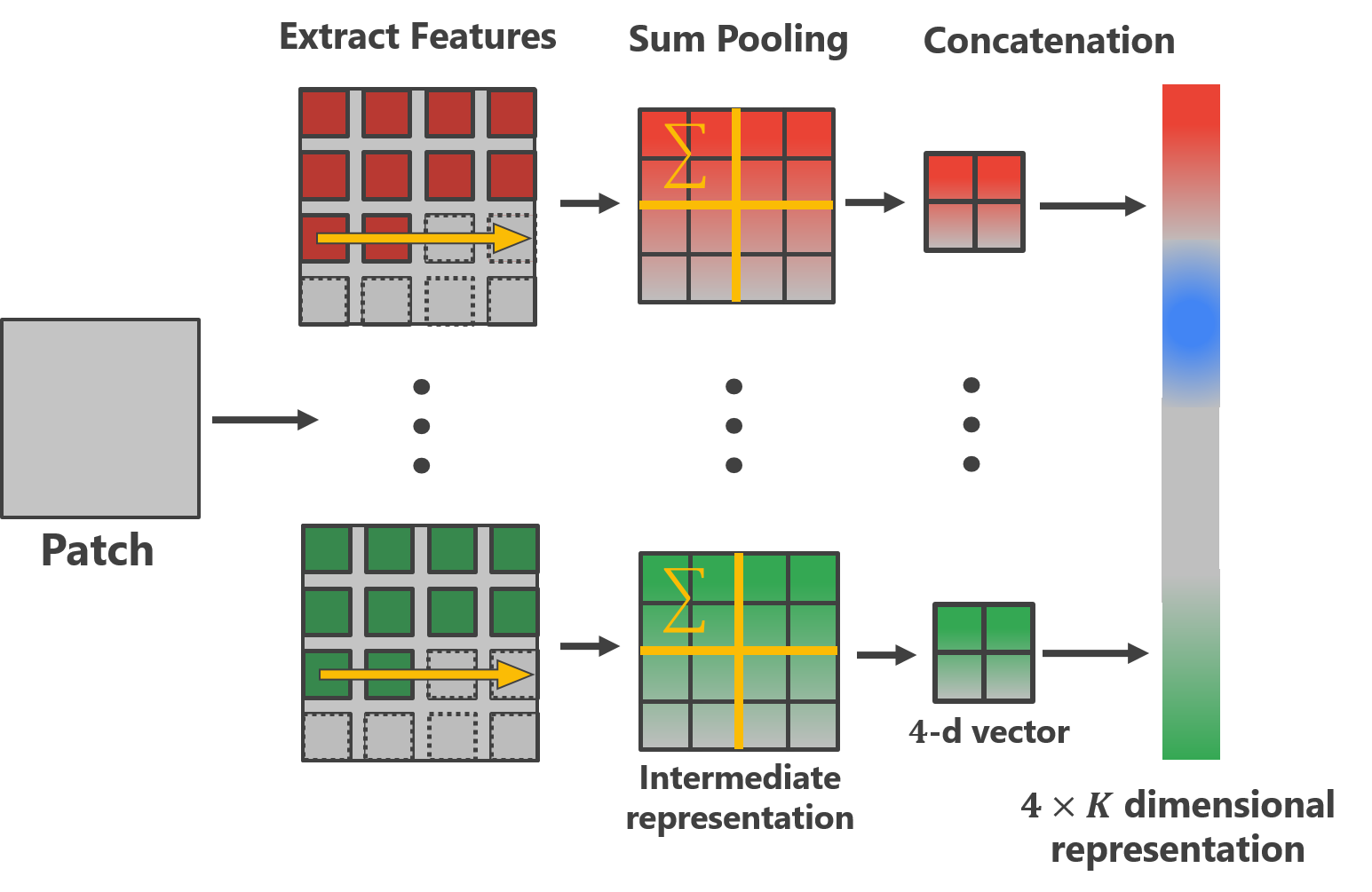}
  \caption{
  Illustration of a pipeline for creating the feature representation.
  We first applied $K$ trained filters in an input patch and obtain $K$ intermediate representations.
  Next, we divide each intermediate representation into 4 equal-sized squares by sum pooling.
  Finally, we concatenate them into an $4 \times K$ dimensional vector used as the final representation.
  }
  \label{fig:extraction}
\end{figure}

%
In the segmentation phase,
we first extract a possible number of patches of $w \times w$  pixels from the target image separated by $s$ pixels each.
Note that stride $s$ is not larger than $w$ in order to ensure overlapping patches.
As with extracting training patches,
we select only voxels within the sample by thresholding.
For transforming image patches to feature representations, we utilize a typical pipeline similar to a single-layered CNN.
Trained filters of $p \times p$ pixels in a patch were applied with a stride of $h$ pixels in order to extract features.
We adopt the soft-threshold nonlinearity as feature extraction function
$f({x};\hat{\mathcal{D}}) = \mathrm{max}\{0, \hat{\mathcal{D}}^{T}{x}\}$.
After feature extraction using $K$ filters,
we obtain an array of $((w-p)/h+1) \times ((w-p)/h+1)$
composed of $K$ intermediate representations.
We reduce the dimensionality of each representation by sum pooling.
Concretely,
we divide each intermediate representation into
four equal-sized squares, and apply sum pooling to each square region
in order to obtain a 4 dimensional pooled vector.
We concatenate pooled vectors from the all intermediate representation into a $4 \times K$ dimensional vector
which is used as the final representation.
The process of feature extraction is illustrated in Fig. \ref{fig:extraction}.
Next, we divide the final representations into $C$ clusters by conventional $k$-means
in order to assign a label $l (1 \le l \le C)$ to each representation.
Finally, we project each labels onto a subpatch of $s\times s$ pixels centered in a corresponding extracted patch.

\section{Experiments and RESULTS}
\label{sec:results}
\subsection{Datasets}
We utilized a set of 70 pathology images from the same lung cancer specimen.
The original size of the images is approximately 200,000$\times$100,000 pixels and
the resolution of the images is 0.220 $\times$0.220 $\mu$m/pixel.
For experiments, we converted it into reduced scale images of approximately 2,000$\times$1,000 pixels and 22$\times$22 $\mu$m/pixel.
The goal of segmentation was to divide each image into three histopathological regions:
(a) invasive carcinoma;
(b) noninvasive carcinoma;
and (c) normal tissue.


\subsection{Parameter Settings}
We prepared 100,000 patches of size $5 \times 5$ pixels randomly extracted from one representative image.
For representation learning with spherical $k$-means, we set the number of clusters $K$ to 200.
In the beginning of segmentation phase, we extracted patches of $99\times99$ pixels with a stride of 1 pixel from a target image.
For creating feature representations, we applied trained filters with a stride of 2 pixels to in order to obtain 800 dimensional feature representations.
For segmentation, we conducted the conventional $k$-means to divide representations into three regions.

\subsection{Evaluations}
\begin{figure}[t]
  \begin{center}
  \includegraphics[keepaspectratio, scale=0.299]{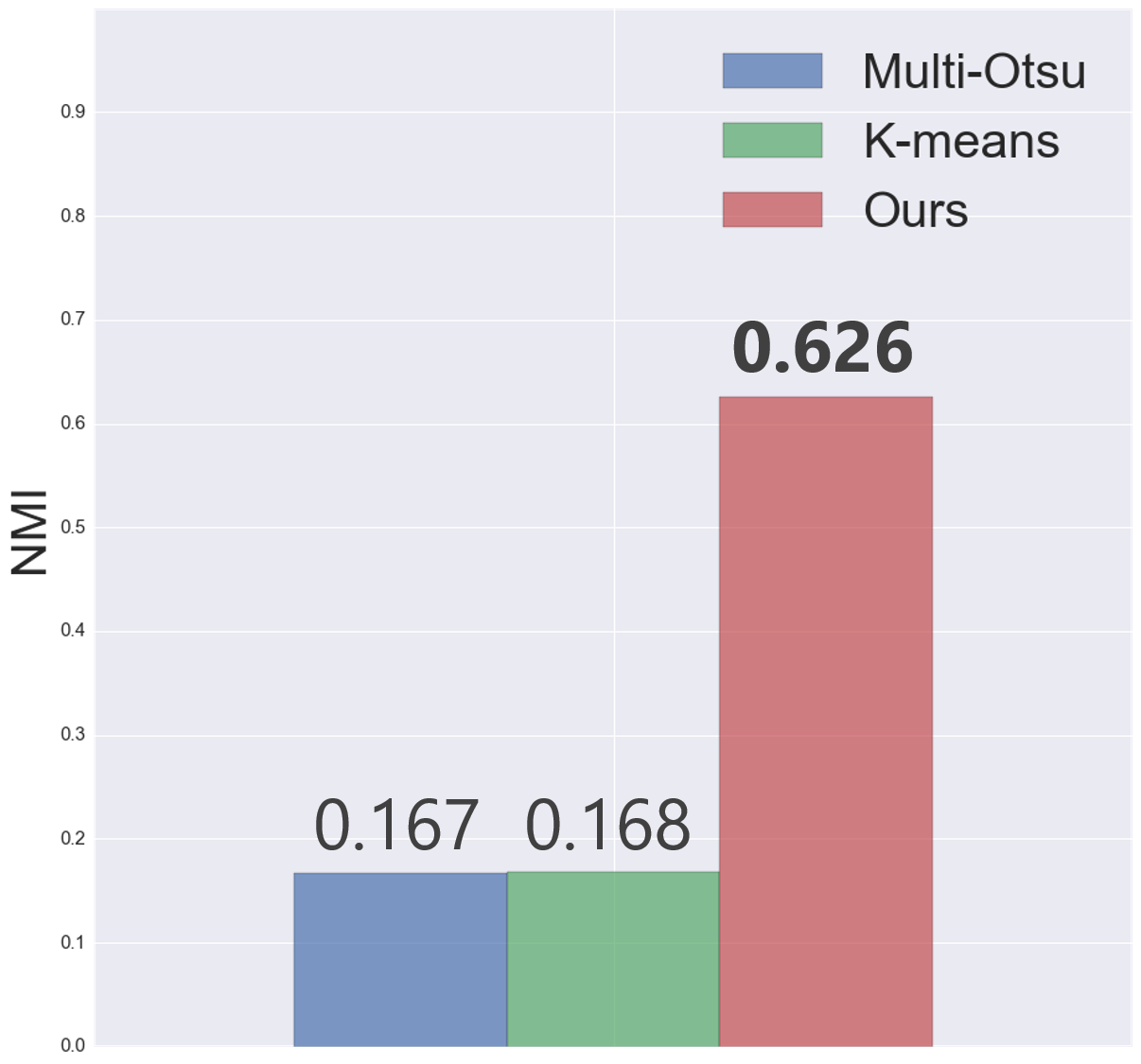}
  \caption{
  Comparison of NMI scores. Score of our method outperforms traditional methods.}
  \label{fig:results}
  \end{center}
\end{figure}

\begin{figure}[tb]
 \begin{minipage}[t]{0.5\linewidth}
   \centering
   \subfloat[Original slice]{\includegraphics[keepaspectratio, scale=0.33]
   {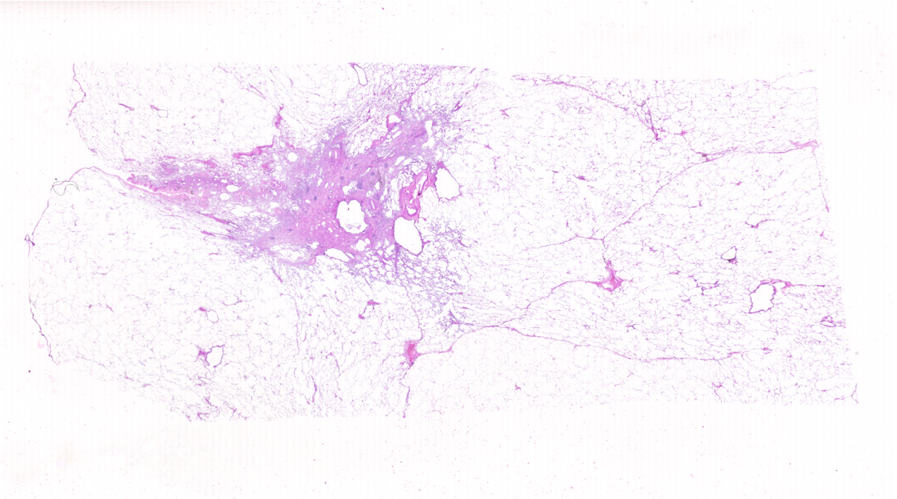}
  }
 \end{minipage}
 \begin{minipage}[t]{0.5\linewidth}
   \centering
   \subfloat[Ground truth]{\includegraphics[keepaspectratio, scale=0.33]
   {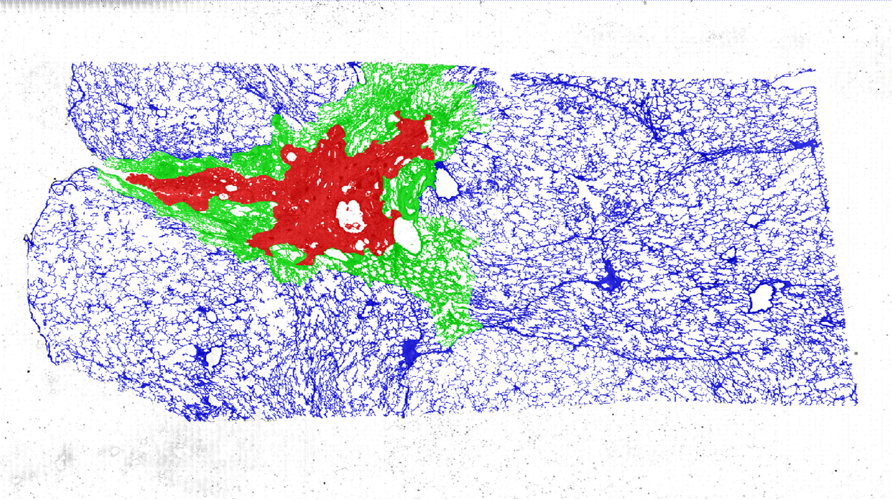}
  }
  \vspace{2em}
 \end{minipage}\\
 \begin{minipage}[t]{0.5\linewidth}
   \centering
   \subfloat[Multi Otsu]{\includegraphics[keepaspectratio, scale=0.33]
   {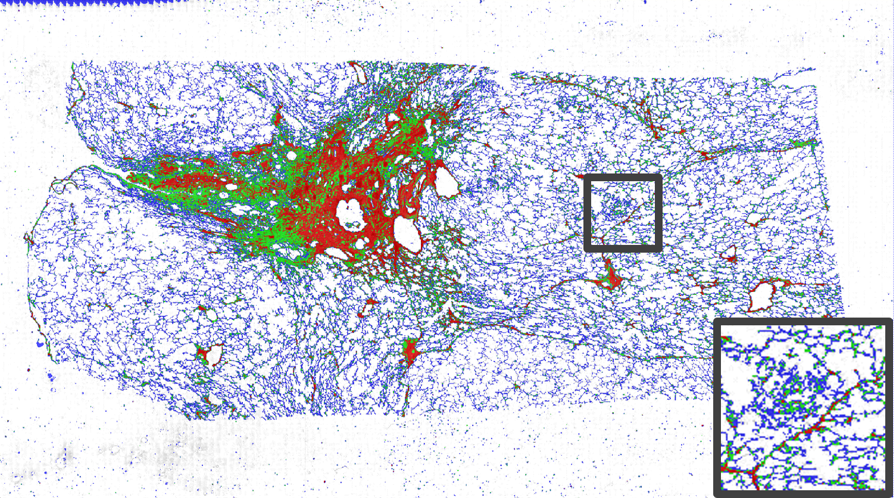}
  }
 \end{minipage}
 \begin{minipage}[t]{0.5\linewidth}
   \centering
   \subfloat[$k$-means]{\includegraphics[keepaspectratio, scale=0.33]
   {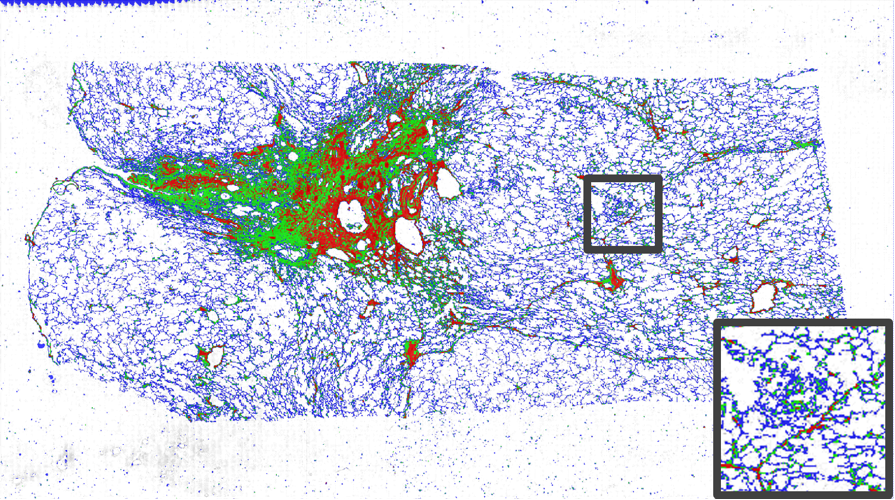}
  }
  \vspace{2em}
 \end{minipage}\\
 \begin{minipage}[t]{\linewidth}
   \centering
   \subfloat[Our result]{\includegraphics[keepaspectratio, scale=0.33]
   {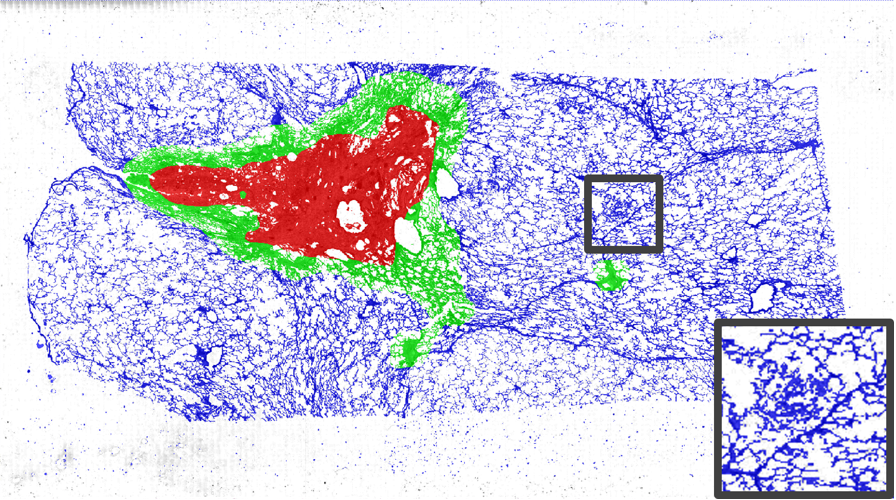}
  }
  \vspace{1em}
 \end{minipage}
   \caption{
   Segmentation result of the pathology image.
   In the ground truth, the red, green, and blue regions correspond to the region of invasive carcinoma, noninvasive carcinoma, and normal tissue, respectively.
   Our method divided pathology image into invasive carcinoma, noninvasive carcinoma, and normal lung better than multithreshold Otsu and conventional $k$-means.
   The lower right images in (c), (d) and (e) is the zoomed region in the black window.
   As shown in the zoomed image, our method causes much less false labels in small structure than multithreshold Otsu and $k$-means.
   }
 \label{fig:result_images}
\end{figure}

\begin{figure}[tb]
  \centering
  \subfloat[3D rendering of Otsu's segmentation results]{\includegraphics[keepaspectratio, scale=0.25]
   {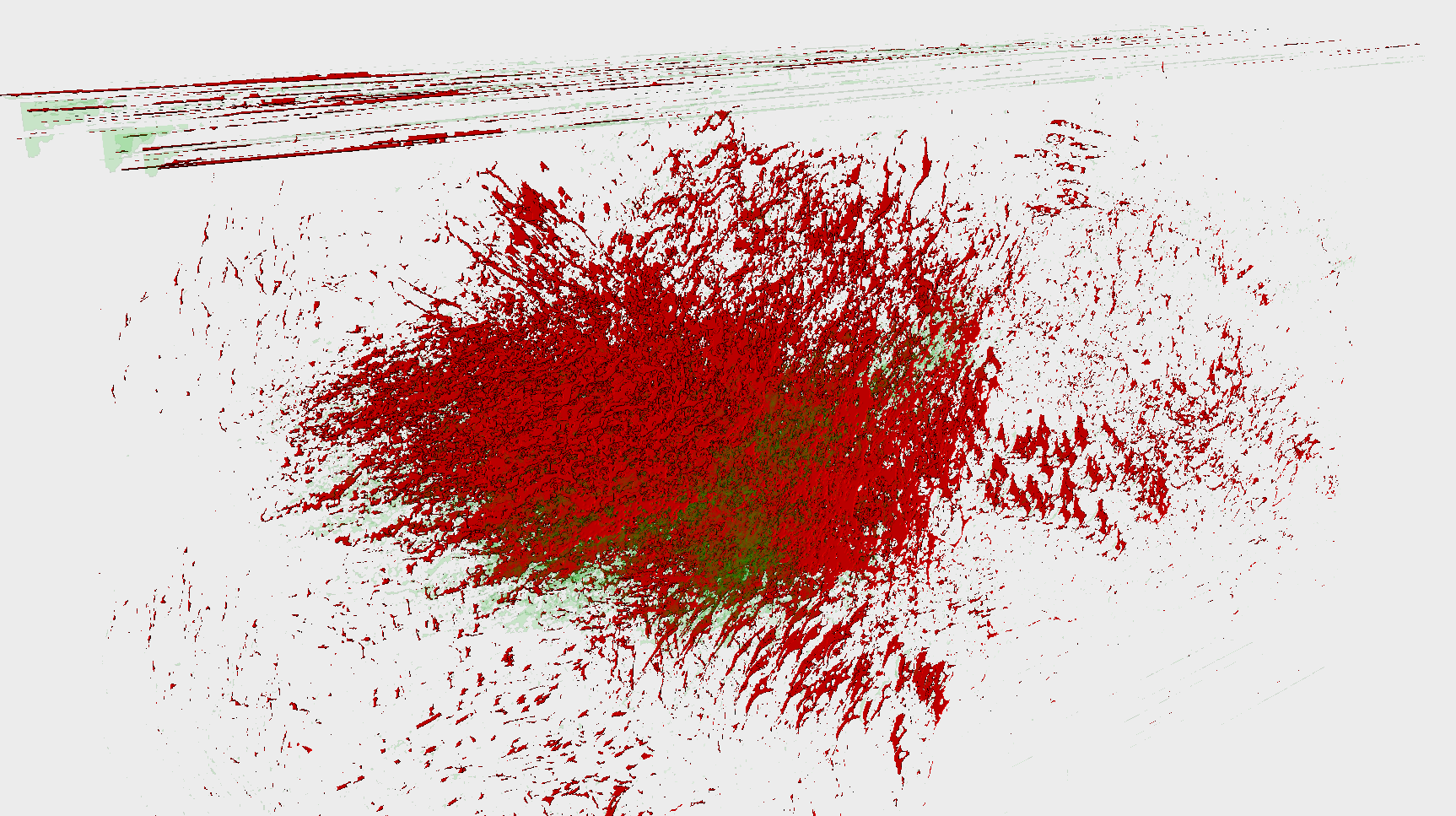}
  }\\
  \centering
  \subfloat[3D rendering of our segmentation results]{\includegraphics[keepaspectratio, scale=0.25]
   {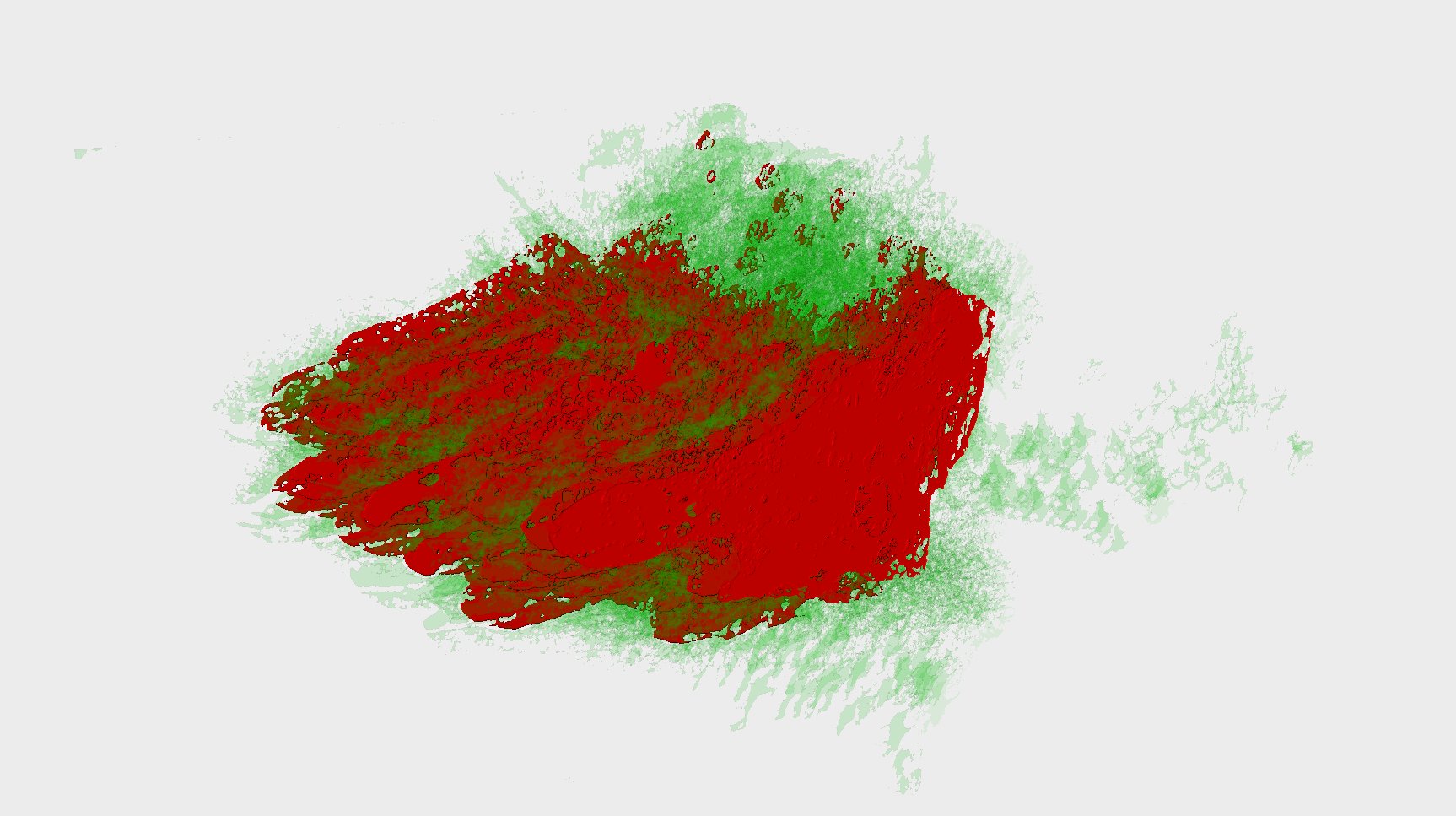}
  }
  \caption{
   3D renderings of segmentation results of all the 70 slices.
   We obtained centroids from one representative and applied them to remaining 69 images for feature extraction and segmentation.
   We only visualize invasive carcinoma (red) and noninvasive carcinoma (semitransparent green).
  }
 \label{fig:result_3d}
\end{figure}

We used one manually annotated image to evaluate the proposed method.
For quantitative evaluation, we used the standard metric for clustering, normalized mutual information (NMI).
A larger NMI value means better segmentation results.
We compared our method with traditional $k$-means segmentation and multithreshold Otsu method \cite{otsu1979threshold}.
As shown in Fig. \ref{fig:results}, our method outperforms traditional methods.
Figure. \ref{fig:result_images} shows a qualitative example produced by the proposed methods.
Our method divided pathology image into invasive carcinoma, noninvasive carcinoma, and normal lung more accurately than Multi Otsu and $k$-means.
Additionally, we applied centroids from one representative slice to remaining 69 slices for feature extraction and segmentation.
Figure. \ref{fig:result_3d} shows 3D renderings of 70 segmented slices.
A rendering of our results is much easier to observe anatomical regions than a rendering of multithreshold Otsu's segmentation results.

\section{Discussions}
Our methods significantly outperformed traditional unsupervised methods both quantitatively and qualitatively.
The reason for this may be that our method could learn not only pixel intensities but also textures of local regions.
Moreover, we found that the centroids obtained by clustering one representative slice can be utilized to extract representations of other slices for segmentation.
It can be suggested that the centroids are not over-fitting to a single slice.
However, our results often caused false labels in higher intensity pixels in normal tissue region.
This is because, seemingly, our method reflected pixel intensities too much.

\section{CONCLUSION}
We proposed a novel unsupervised segmentation method that obtains segmented images by clustering feature representations.
Our proposed method outperforms the traditional unsupervised methods.
We demonstrated the potential abilities of unsupervised representation learning for
pathology image segmentation.
Our segmentation method could be applicable to both 2D and 3D medical imaging applications.


%
%
\section*{ACKNOWLEDGMENTS}
This research was supported by the Kakenhi by MEXT and JSPS (26108006, 17K20099) and the JSPS Bilateral International Collaboration Grants.

\clearpage
\bibliography{report} 
\bibliographystyle{spiebib} 

\end{document}